\documentclass[11pt]{article}
\usepackage{acl}
\usepackage{times}
\usepackage{latexsym}
\usepackage[T1]{fontenc}
\usepackage[utf8]{inputenc}
\usepackage{microtype}
\usepackage{inconsolata}
\usepackage{booktabs}
\usepackage{graphicx}
\usepackage{amsmath}
\usepackage{multirow}
\usepackage{xcolor}
\usepackage{subcaption}
\graphicspath{{./figures/}}

\title{Thinking Fast, Thinking Wrong: Intuitiveness Modulates\\LLM Counterfactual Reasoning in Policy Evaluation}

\author{Yanjie He \\
  Independent Researcher \\
  \texttt{heyanjie0@outlook.com}}

\begin{document}
% Auto-generated by generate_latex_stats.py — DO NOT EDIT MANUALLY
% Source: paper_stats.json
% Run tag: full_v3

% ===== Overall Descriptive Statistics =====
\newcommand{\statNtrials}{8000}
\newcommand{\statNcorrect}{6369}
\newcommand{\statAccOverall}{79.6}
\newcommand{\statNcases}{40}

% ===== Case Distribution (Table 1) =====
\newcommand{\statNobviousOneSided}{0}
\newcommand{\statNobviousTwoSided}{13}
\newcommand{\statNambiguousOneSided}{1}
\newcommand{\statNambiguousTwoSided}{5}
\newcommand{\statNciOneSided}{13}
\newcommand{\statNciTwoSided}{8}
\newcommand{\statNoneSided}{14}
\newcommand{\statNtwoSided}{26}
\newcommand{\statNci}{21}

% ===== By Test Type =====
\newcommand{\statAccOneSided}{98.7}
\newcommand{\statAccTwoSided}{69.3}

% ===== Two-Sided Accuracy by Intuitiveness =====
\newcommand{\statAccTsObvious}{68.2}
\newcommand{\statAccTsAmbiguous}{72.9}
\newcommand{\statAccTsCI}{69}

% ===== Model Accuracy =====
\newcommand{\statAccModelGptFive}{96.3}
\newcommand{\statAccModelGptFiveTs}{94.3}
\newcommand{\statAccModelGptFiveTsCI}{94.5}
\newcommand{\statAccModelSonnet}{79.8}
\newcommand{\statAccModelSonnetTs}{69.6}
\newcommand{\statAccModelSonnetTsCI}{73.5}
\newcommand{\statAccModelOpus}{74.7}
\newcommand{\statAccModelOpusTs}{61.7}
\newcommand{\statAccModelOpusTsCI}{64.8}
\newcommand{\statAccModelGptFour}{67.6}
\newcommand{\statAccModelGptFourTs}{51.7}
\newcommand{\statAccModelGptFourTsCI}{43.2}

% ===== Prompt Strategy Accuracy =====
\newcommand{\statAccPromptNaive}{80.8}
\newcommand{\statAccPromptNaiveTs}{72}
\newcommand{\statAccPromptNaiveTsCI}{70}
\newcommand{\statAccPromptExpert}{68.6}
\newcommand{\statAccPromptExpertTs}{53.3}
\newcommand{\statAccPromptExpertTsCI}{64.7}
\newcommand{\statAccPromptCot}{95.5}
\newcommand{\statAccPromptCotTs}{93.1}
\newcommand{\statAccPromptCotTsCI}{83.8}
\newcommand{\statAccPromptStruct}{96.1}
\newcommand{\statAccPromptStructTs}{94}
\newcommand{\statAccPromptStructTsCI}{85}
\newcommand{\statAccPromptAdv}{57.1}
\newcommand{\statAccPromptAdvTs}{34.2}
\newcommand{\statAccPromptAdvTsCI}{41.6}

% ===== Majority Vote =====
\newcommand{\statAccMV}{80.5}
\newcommand{\statAccMVTsCI}{71.9}

% ===== Repetition Consistency =====
\newcommand{\statRepConsistency}{79.5}

% ===== Hardest Cases =====
\newcommand{\statAccCaseRTC}{38}
\newcommand{\statAccCaseDaycare}{39}

% ===== Error Patterns =====
\newcommand{\statErrReversalPct}{72}
\newcommand{\statErrPTwoNscPct}{65}

% ===== CoT Gain (Two-Sided) =====
\newcommand{\statCotGainObvious}{26.7}
\newcommand{\statCotGainAmbiguous}{18}
\newcommand{\statCotGainCI}{13.8}

% ===== Regression: Main Effects =====
\newcommand{\statORcotMain}{17.3}
\newcommand{\statCIcotMain}{10.1, 29.4}
\newcommand{\statPcotMain}{{$<$0.001}}

% ===== Regression: Selected Coefficients =====
\newcommand{\statORstructMain}{22.6}
\newcommand{\statCIstructMain}{12.6, 40.7}
\newcommand{\statPstructMain}{{$<$0.001}}
\newcommand{\statORadvMain}{0.089}
\newcommand{\statCIadvMain}{0.063, 0.125}
\newcommand{\statPadvMain}{{$<$0.001}}
\newcommand{\statORexpertMain}{0.223}
\newcommand{\statCIexpertMain}{0.164, 0.304}
\newcommand{\statPexpertMain}{{$<$0.001}}
\newcommand{\statORmodelGptFour}{0.013}
\newcommand{\statCImodelGptFour}{0.009, 0.019}
\newcommand{\statPmodelGptFour}{{$<$0.001}}
\newcommand{\statORmodelOpus}{0.030}
\newcommand{\statCImodelOpus}{0.022, 0.042}
\newcommand{\statPmodelOpus}{{$<$0.001}}
\newcommand{\statORmodelSonnet}{0.058}
\newcommand{\statCImodelSonnet}{0.042, 0.079}
\newcommand{\statPmodelSonnet}{{$<$0.001}}
\newcommand{\statORintuitCI}{36.2}
\newcommand{\statCIintuitCI}{5.51, 237.7}
\newcommand{\statPintuitCI}{{$<$0.001}}

% ===== Regression: Interactions (Table 3) =====
\newcommand{\statORCotXCI}{0.278}
\newcommand{\statCICotXCI}{0.140, 0.560}
\newcommand{\statPCotXCI}{{$<$0.001}}
\newcommand{\statORStructXCI}{0.244}
\newcommand{\statCIStructXCI}{0.110, 0.520}
\newcommand{\statPStructXCI}{{$<$0.001}}
\newcommand{\statORAdvXCI}{2.83}
\newcommand{\statCIAdvXCI}{1.70, 4.71}
\newcommand{\statPAdvXCI}{{$<$0.001}}

% ===== Random Effects =====
\newcommand{\statSigmaSqU}{6.706}
\newcommand{\statICC}{0.671}

% ===== Familiarity Moderator =====
\newcommand{\statORfam}{1.08}
\newcommand{\statCIfam}{0.510, 2.27}
\newcommand{\statPfam}{{0.84}}

% ===== Derived: CoT Attenuation =====
\newcommand{\statEffectiveORCotCI}{4.8}
\newcommand{\statLnORcotMain}{2.85}
\newcommand{\statLnEffectiveORCotCI}{1.57}
\newcommand{\statAttenuationPct}{45}

% ===== Robustness: Model × Intuitiveness =====
\newcommand{\statRobORCotXCI}{0.331}
\newcommand{\statRobPCotXCI}{{0.003}}

\maketitle

\begin{abstract}
Large language models (LLMs) are increasingly used for causal and counterfactual reasoning, yet their reliability in real-world policy evaluation remains underexplored. We construct a benchmark of \statNcases{} empirical policy evaluation cases drawn from economics and social science, each grounded in peer-reviewed evidence and classified by \emph{intuitiveness}---whether the empirical finding aligns with (obvious), is unclear relative to (ambiguous), or contradicts (counter-intuitive) common prior expectations. We evaluate four frontier LLMs across five prompting strategies with \statNtrials{} experimental trials and analyze the results using mixed-effects logistic regression. Our findings reveal three key results: (1) a \textbf{chain-of-thought (CoT) paradox}, where chain-of-thought prompting dramatically improves performance on obvious cases but this benefit \emph{is substantially attenuated} on counter-intuitive ones (interaction OR = \statORCotXCI{}, $p$ \statPCotXCI{}); (2) \textbf{intuitiveness as the dominant factor}, with case-level variance exceeding that of model choice or prompting strategy (ICC = \statICC{}); and (3) a \textbf{knowledge--reasoning dissociation}, where citation-based familiarity is unrelated to accuracy ($p =$ \statPfam{}), suggesting models possess relevant knowledge but fail to reason with it when findings contradict intuition. We frame these results through the lens of dual-process theory (System~1 vs.\ System~2) and argue that current LLMs' ``slow thinking'' achieves only partial inhibition of intuitive priors---producing the form of deliberative reasoning without fully delivering its substance.
\end{abstract}

% ============================================================
\section{Introduction}
\label{sec:intro}
% ============================================================

Large language models (LLMs) have demonstrated impressive capabilities across a wide range of reasoning tasks \citep{brown2020language, openai2023gpt4}. Recent work suggests they can perform causal reasoning with high accuracy, surpassing existing methods on standard benchmarks \citep{kiciman2023causal}. These findings have fueled optimism about using LLMs as assistants for causal inference and policy analysis.

However, most existing evaluations rely on synthetic causal graphs \citep{jin2023cladder}, commonsense scenarios, or abstract logical problems \citep{huangACL2024clomo}. The real-world domain of \emph{policy evaluation}---where economists and social scientists use quasi-experimental methods to estimate causal effects---presents a fundamentally different challenge. Policy effects are messy, contested, and, critically, sometimes \emph{counter-intuitive}: the data show the opposite of what most people would expect.

We introduce a benchmark of \statNcases{} empirical policy evaluation cases drawn from peer-reviewed economics and social science research. Each case is classified along an \textbf{intuitiveness} dimension:
\begin{itemize}
    \item \textbf{Obvious}: The empirical result aligns with common prior expectations (e.g., Medicaid expansion increases healthcare utilization).
    \item \textbf{Ambiguous}: Reasonable arguments exist for multiple directions (e.g., welfare reform may increase or decrease employment depending on the mechanism).
    \item \textbf{Counter-intuitive}: The empirical result contradicts na\"ive expectations (e.g., introducing fines for late daycare pickup \emph{increases} lateness).
\end{itemize}

This classification is inspired by dual-process theory from cognitive psychology \citep{kahneman2011thinking, stanovich2000individual, evans2003two}: obvious cases can be solved by fast, intuitive ``System~1'' processing, while counter-intuitive cases require slow, deliberative ``System~2'' reasoning that overrides the initial heuristic response.

We evaluate four frontier LLMs---GPT-5.2, Claude Sonnet~4.5, Claude Opus~4.6, and GPT-4.1---across five prompting strategies (na\"ive, expert persona, chain-of-thought, structured reasoning, and adversarial framing) with 10 repetitions per condition, yielding \statNtrials{} experimental trials. We analyze the results using mixed-effects logistic regression to properly account for the nested structure of the data.

Our key findings are: (1) a \textbf{CoT paradox}---chain-of-thought prompting substantially improves accuracy on obvious cases (main-effect OR = \statORcotMain{}) but this benefit \emph{is attenuated} on counter-intuitive ones (interaction OR = \statORCotXCI{}, $p$ \statPCotXCI{}); (2) \textbf{intuitiveness as the dominant factor}, with case-level random intercepts capturing the majority of variance (ICC = \statICC{}); and (3) a \textbf{knowledge--reasoning dissociation}, where citation-based familiarity is unrelated to accuracy ($p =$ \statPfam{}), suggesting models possess relevant knowledge but fail to reason with it when findings contradict intuition. Overall accuracy of \statAccOverall{}\% masks critical failures precisely where counterfactual reasoning matters most.

% ============================================================
\section{Related Work}
\label{sec:related}
% ============================================================

\paragraph{Causal Reasoning in LLMs.}
\citet{kiciman2023causal} report 92\% accuracy on counterfactual reasoning but rely on commonsense and synthetic scenarios. \citet{jin2023cladder} propose CLadder using synthetic causal graphs with oracle-derived ground truth. \citet{zevcevic2023causal} argue LLMs are ``causal parrots'' reciting memorized knowledge. Our work bridges these perspectives: LLMs \emph{can} leverage causal knowledge, but fail when doing so requires \emph{overriding} intuitive priors---a distinction invisible in synthetic benchmarks.

\paragraph{Counterfactual Reasoning Benchmarks.}
\citet{huangACL2024clomo} propose CLOMO for generative counterfactual modification; \citet{muACL2024causal} use VAE-based counterfactual reasoning in narratives; \citet{weinzierlACL2024tree} develop Tree-of-Counterfactual prompting. Unlike these works, we focus on \emph{evaluative} counterfactual reasoning in real-world policy evaluation, where ground truth is established by peer-reviewed empirical research.

\paragraph{Cognitive Parallels in LLM Behavior.}
\citet{yingACL2024intuitive} classify LLMs' response styles as ``intuitive'' or ``dependent''; their ``intuitive'' refers to behavioral style, whereas ours refers to cognitive accessibility of task content. \citet{mondorfACL2024comparing} find that LLM accuracy does not necessarily reflect reasoning validity---a conclusion our findings reinforce. \citet{shenAAAI2025stressprompt} show that LLMs follow the Yerkes-Dodson law under stress-inducing prompts.

\paragraph{Chain-of-Thought Reasoning.}
CoT prompting \citep{weiNeurIPS2022cot, kojima2022large} is widely assumed to improve reasoning, but \citet{jacoviACL2024cot} show that reasoning chains are only as strong as their weakest step. Our CoT paradox extends this: the weakest step may be the very first one---the intuitive prior that anchors subsequent reasoning. \citet{snellICLR2025scaling} show that test-time compute scaling depends critically on prompt difficulty, consistent with our finding.

% ============================================================
\section{Benchmark Design}
\label{sec:benchmark}
% ============================================================

\subsection{Case Selection}
\label{sec:cases}

We curate \statNcases{} empirical policy evaluation cases from economics/social science, each satisfying three criteria: (1) at least one peer-reviewed publication establishing the causal effect via a credible identification strategy (difference-in-differences, regression discontinuity, instrumental variables, or randomized controlled trial); (2) an unambiguous directional finding (increase, decrease, or no significant change); and (3) sufficient prominence that the study is likely present in LLM training data. Cases span 11 countries, 10 policy domains, and publication years from 1990 to 2020. Table~\ref{tab:cases_summary} summarizes the distribution.

\begin{table}[t]
\centering
\small
\begin{tabular}{lrr}
\toprule
\textbf{Intuitiveness} & \textbf{One-sided} & \textbf{Two-sided} \\
\midrule
Obvious      & \statNobviousOneSided{}  & \statNobviousTwoSided{} \\
Ambiguous    & \statNambiguousOneSided{}  & \statNambiguousTwoSided{}  \\
Counter-intuitive & \statNciOneSided{} & \statNciTwoSided{}  \\
\midrule
\textbf{Total}    & \statNoneSided{} & \statNtwoSided{} \\
\bottomrule
\end{tabular}
\caption{Distribution of \statNcases{} benchmark cases by intuitiveness and test type. One-sided cases ask a YES/NO question; two-sided cases ask for directional prediction (INCREASE / DECREASE / NO SIGNIFICANT CHANGE).}
\label{tab:cases_summary}
\end{table}

\subsection{Intuitiveness Classification}
\label{sec:intuitiveness}

Each case is classified as \textbf{obvious}, \textbf{ambiguous}, or \textbf{counter-intuitive} based on whether the established empirical finding aligns with, is unclear relative to, or contradicts the prior that an informed layperson would hold. This classification was performed by the author and validated by three independent human annotators (Fleiss' $\kappa$ = 0.32, indicating fair agreement; see Appendix~\ref{app:annotation}). The categorical classification is further supported by its convergent validity with the experimental results: CoT prompting yields large gains on obvious cases (+\statCotGainObvious{} pp) but substantially attenuated gains on counter-intuitive cases (+\statCotGainCI{} pp), consistent with the dual-process interpretation (\S\ref{sec:cot_paradox}).

Examples of counter-intuitive cases include:
\begin{itemize}
    \item Introducing fines for late daycare pickup \emph{increases} lateness \citep{gneezyRustichini2000}: monetary penalties replace moral obligation with a market transaction.
    \item ``Right-to-carry'' concealed handgun laws \emph{increase} violent crime \citep{donohueEtAl2019}: contrary to deterrence theory.
    \item Violent movie releases \emph{decrease} violent crime \citep{dahlDellavigna2009}: voluntary incapacitation outweighs any media-effects priming.
\end{itemize}

\subsection{Test Types}
\label{sec:test_types}

We employ two test types. \textbf{One-sided} tests present a directional hypothesis and ask the model to answer YES or NO (e.g., ``Did the minimum wage increase decrease employment? YES or NO''). \textbf{Two-sided} tests ask the model to predict the direction of the counterfactual (INCREASE / DECREASE / NO SIGNIFICANT CHANGE). Two-sided tests are strictly harder, as the model must both identify whether an effect exists and determine its sign.

\subsection{Prompt Design}
\label{sec:prompts}

We design five prompting strategies to probe different aspects of reasoning:

\begin{enumerate}
    \item \textbf{P1 -- Na\"ive:} A direct question with no guidance, capturing the model's default ``System~1'' response.
    \item \textbf{P2 -- Expert Persona:} The model is told it is an ``experienced econometrician'' to test whether domain framing improves accuracy.
    \item \textbf{P3 -- Chain-of-Thought (CoT):} The model is asked to ``think step by step,'' listing causal mechanisms and confounders before answering.
    \item \textbf{P4 -- Structured Reasoning:} A more detailed template requiring explicit enumeration of causal channels, confounders, and finally a prediction.
    \item \textbf{P5 -- Adversarial Framing:} The prompt introduces a fictitious ``analyst consensus'' suggesting the policy had no effect, testing epistemic robustness against anchoring bias \citep{tversky1974judgment}.
\end{enumerate}

% ============================================================
\section{Experimental Setup}
\label{sec:setup}
% ============================================================

\subsection{Models}

We evaluate four frontier LLMs accessed via API:

\begin{itemize}
    \item \textbf{GPT-5.2} (OpenAI)
    \item \textbf{Claude Sonnet 4.5} (Anthropic)
    \item \textbf{Claude Opus 4.6} (Anthropic)
    \item \textbf{GPT-4.1} (OpenAI)
\end{itemize}

All models are accessed via default inference parameters, held constant across all experimental conditions. To prevent data contamination during evaluation, we sandbox each call: file-system tools are disabled, the working directory is set to a temporary folder, and custom instructions are suppressed.

\subsection{Procedure}

Each of the $\statNcases{} \times 4 \times 5 \times 10 = \statNtrials{}$ trials is executed independently. Responses are parsed using regular expressions to extract the final answer (INCREASE / DECREASE / NO SIGNIFICANT CHANGE / YES / NO), with a lightweight LLM fallback parser for the $<1\%$ of responses where regex extraction fails.

\subsection{Statistical Analysis}

We analyze the binary outcome (correct/incorrect) using mixed-effects logistic regression \citep{bates2015lme4}:

\begin{equation}
\label{eq:model}
\text{logit}(P(\text{correct}_{ij})) = \mathbf{X}_{ij}\boldsymbol{\beta} + u_j
\end{equation}

\noindent where $i$ indexes observations and $j$ indexes cases. The fixed effects $\boldsymbol{\beta}$ include model identity, prompt strategy, intuitiveness category, and their interactions (prompt $\times$ intuitiveness). The random intercept $u_j \sim \mathcal{N}(0, \sigma_u^2)$ captures case-level heterogeneity. This specification follows recommendations in \citet{barr2013random} for confirmatory hypothesis testing. We report odds ratios (OR) with 95\% confidence intervals. All analyses are implemented in R using \texttt{lme4} \citep{bates2015lme4}.

% ============================================================
\section{Results}
\label{sec:results}
% ============================================================

\subsection{Overall Performance}

Across all \statNtrials{} trials, overall accuracy is \statAccOverall{}\% (\statNcorrect{}/\statNtrials{}). However, this aggregate masks substantial heterogeneity. One-sided tests yield \statAccOneSided{}\% accuracy (near ceiling), while two-sided tests yield only \statAccTwoSided{}\%. Within two-sided tests, accuracy is uniformly low across intuitiveness categories (obvious \statAccTsObvious{}\%, counter-intuitive \statAccTsCI{}\%, ambiguous \statAccTsAmbiguous{}\%), suggesting that the challenge lies not in raw accuracy differences but in how prompting strategies interact with intuitiveness (\S\ref{sec:cot_paradox}).

\subsection{Model Comparison}

\begin{table}[t]
\centering
\small
\begin{tabular}{lrrr}
\toprule
\textbf{Model} & \textbf{Overall} & \textbf{Two-sided} & \textbf{TS c-intuit.} \\
\midrule
GPT-5.2        & \statAccModelGptFive{}\% & \statAccModelGptFiveTs{}\% & \statAccModelGptFiveTsCI{}\% \\
Claude Sonnet 4.5 & \statAccModelSonnet{}\% & \statAccModelSonnetTs{}\% & \statAccModelSonnetTsCI{}\% \\
Claude Opus 4.6 & \statAccModelOpus{}\% & \statAccModelOpusTs{}\% & \statAccModelOpusTsCI{}\% \\
GPT-4.1         & \statAccModelGptFour{}\% & \statAccModelGptFourTs{}\% & \statAccModelGptFourTsCI{}\% \\
\bottomrule
\end{tabular}
\caption{Accuracy by model. TS c-intuit.\ = two-sided counter-intuitive cases only. GPT-5.2 maintains high performance even on counter-intuitive cases. Claude Sonnet~4.5 outperforms Claude Opus~4.6 on counter-intuitive cases despite being a smaller model, suggesting model scale alone does not determine susceptibility to intuitive priors.}
\label{tab:model_results}
\end{table}

Table~\ref{tab:model_results} shows that GPT-5.2 substantially outperforms all other models, particularly on two-sided counter-intuitive cases, where GPT-4.1 falls to \statAccModelGptFourTsCI{}\%---only 10 percentage points above the 33\% chance baseline. Notably, Claude Sonnet~4.5 outperforms Claude Opus~4.6 on counter-intuitive cases (\statAccModelSonnetTsCI{}\% vs.\ \statAccModelOpusTsCI{}\%) despite being a smaller model, suggesting that model scale alone does not determine susceptibility to intuitive priors (see also Figure~\ref{fig:model_intuitiveness} in Appendix~\ref{app:additional}).

\subsection{Prompt Strategy Comparison}

\begin{table}[t]
\centering
\small
\begin{tabular}{lrrr}
\toprule
\textbf{Prompt} & \textbf{Overall} & \textbf{Two-sided} & \textbf{TS c-intuit.} \\
\midrule
P4 Structured   & \statAccPromptStruct{}\% & \statAccPromptStructTs{}\% & \statAccPromptStructTsCI{}\% \\
P3 CoT          & \statAccPromptCot{}\% & \statAccPromptCotTs{}\% & \statAccPromptCotTsCI{}\% \\
P1 Na\"ive      & \statAccPromptNaive{}\% & \statAccPromptNaiveTs{}\% & \statAccPromptNaiveTsCI{}\% \\
P2 Expert       & \statAccPromptExpert{}\% & \statAccPromptExpertTs{}\% & \statAccPromptExpertTsCI{}\% \\
P5 Adversarial  & \statAccPromptAdv{}\% & \statAccPromptAdvTs{}\% & \statAccPromptAdvTsCI{}\% \\
\bottomrule
\end{tabular}
\caption{Accuracy by prompt strategy. The expert persona (P2) underperforms the na\"ive baseline (P1), while the adversarial prompt (P5) severely degrades overall accuracy. CoT and structured prompts remain effective on counter-intuitive cases, though with attenuated benefit (\S\ref{sec:cot_paradox}).}
\label{tab:prompt_results}
\end{table}

Structured reasoning (P4) and CoT (P3) dominate overall, but these aggregates conceal the critical interaction with intuitiveness (\S\ref{sec:cot_paradox}).

\subsection{The CoT Paradox}
\label{sec:cot_paradox}

The central finding of this paper emerges from the interaction between prompting strategy and intuitiveness. Table~\ref{tab:interaction} presents the key interaction terms from our mixed-effects model.

\begin{table}[t]
\centering
\small
\setlength{\tabcolsep}{4pt}
\begin{tabular}{@{}lrrr@{}}
\toprule
\textbf{Interaction} & \textbf{OR} & \textbf{95\% CI} & \textbf{$p$} \\
\midrule
P3 CoT $\times$ c-intuit. & \statORCotXCI{} & [\statCICotXCI{}] & \statPCotXCI{} \\
P4 Struct. $\times$ c-intuit. & \statORStructXCI{} & [\statCIStructXCI{}] & \statPStructXCI{} \\
P5 Advers. $\times$ c-intuit. & \statORAdvXCI{} & [\statCIAdvXCI{}] & \statPAdvXCI{} \\
\midrule
\multicolumn{4}{@{}l}{\textit{Random effects}} \\
$\sigma^2_u$ (case intercept) & \multicolumn{3}{l}{\statSigmaSqU{}} \\
ICC & \multicolumn{3}{l}{\statICC{}} \\
\bottomrule
\end{tabular}
\caption{Key interaction terms from the mixed-effects logistic regression (Eq.~\ref{eq:model}). CoT and structured prompting show a substantial \emph{reduction} of benefit on counter-intuitive cases (OR $<$ 1). The P5 positive interaction ($p$ \statPAdvXCI{}) is confounded by test-type imbalance (see text).}
\label{tab:interaction}
\end{table}

The CoT $\times$ counter-intuitive interaction OR of \statORCotXCI{} means that CoT's large benefit on obvious cases (main-effect OR = \statORcotMain{}) is substantially attenuated on counter-intuitive cases. Structured reasoning shows a similar pattern (OR = \statORStructXCI{}). CoT still helps on counter-intuitive cases (effective OR $\approx$ \statORcotMain{} $\times$ \statORCotXCI{} $\approx$ \statEffectiveORCotCI{}), but the attenuation is substantial ($p$ \statPCotXCI{}).\footnote{A robustness check including model $\times$ intuitiveness interactions yields qualitatively identical results (CoT $\times$ CI: OR = \statRobORCotXCI{}, $p$ \statRobPCotXCI{}), confirming that the attenuation is not driven by model-specific effects.}

The adversarial prompt (P5) shows a significant \emph{positive} interaction with counter-intuitive cases (OR = \statORAdvXCI{}, $p$ \statPAdvXCI{}). However, this result is confounded by test type: \statNciOneSided{} of \statNci{} counter-intuitive cases are one-sided with ground truth NO, and P5's ``no effect'' anchor coincidentally aligns with the correct answer. On two-sided counter-intuitive cases---where the ground truth is INCREASE or DECREASE---P5 \emph{reduces} accuracy from \statAccPromptNaiveTsCI{}\% (P1) to \statAccPromptAdvTsCI{}\%, the worst of any prompt. The apparent positive interaction thus reflects accidental alignment between the anchor and the correct answer, not genuine debiasing.

\begin{figure}[t]
\centering
\includegraphics[width=\columnwidth]{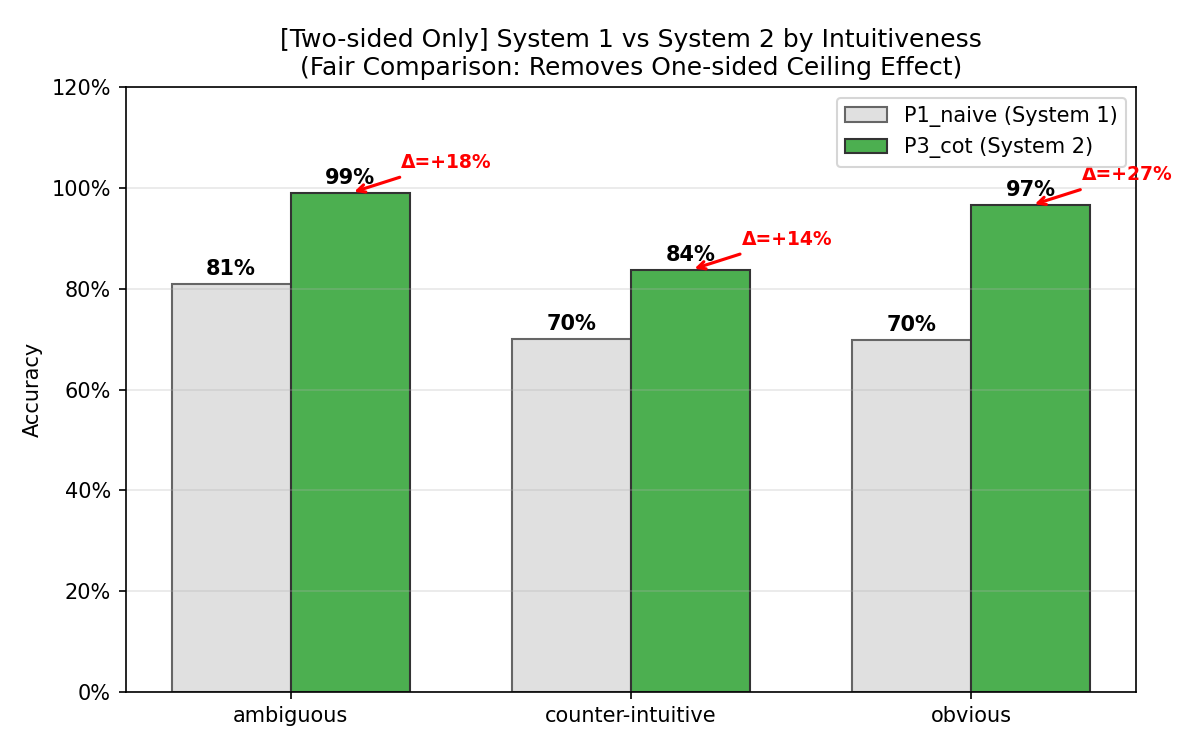}
\caption{The CoT paradox: change in accuracy from na\"ive to CoT prompting by intuitiveness category (two-sided cases only). CoT improves accuracy by +\statCotGainObvious{} percentage points on obvious cases but only +\statCotGainCI{} points on counter-intuitive cases, with this gap driven by CoT partially reinforcing rather than fully overriding intuitive priors.}
\label{fig:cot_paradox}
\end{figure}

Figure~\ref{fig:cot_paradox} illustrates the CoT paradox: the accuracy gain from chain-of-thought prompting diminishes substantially as intuitiveness decreases.

\subsection{Self-Consistency Analysis}

Majority-vote accuracy \citep{wang2023selfconsistency} across ten repetitions improves marginally from \statAccOverall{}\% to \statAccMV{}\% overall, but provides \emph{minimal improvement} on two-sided counter-intuitive cases (\statAccTsCI{}\% $\to$ \statAccMVTsCI{}\%). Errors are \emph{systematic, not stochastic}: models consistently give the same wrong answer, reflecting stable but incorrect causal beliefs.

% ============================================================
\section{Analysis}
\label{sec:analysis}
% ============================================================

\subsection{Error Patterns}

Among incorrect two-sided responses where the ground truth is INCREASE, \statErrReversalPct{}\% of errors predict DECREASE (directional reversal) rather than NO SIGNIFICANT CHANGE---consistent with models applying an intuitive causal model that predicts the \emph{opposite} direction. The expert persona (P2) shows a distinctive pattern: \statErrPTwoNscPct{}\% of its errors are NO SIGNIFICANT CHANGE, suggesting that ``econometrician'' framing induces excessive hedging.

\subsection{Case Difficulty}

The random intercepts ($u_j$) reveal substantial case-level heterogeneity. The hardest cases are Case~23 (daycare fines; \statAccCaseDaycare{}\%) and Case~36 (right-to-carry laws; \statAccCaseRTC{}\%), both counter-intuitive and two-sided.

\begin{figure*}[t]
\centering
\includegraphics[width=\textwidth]{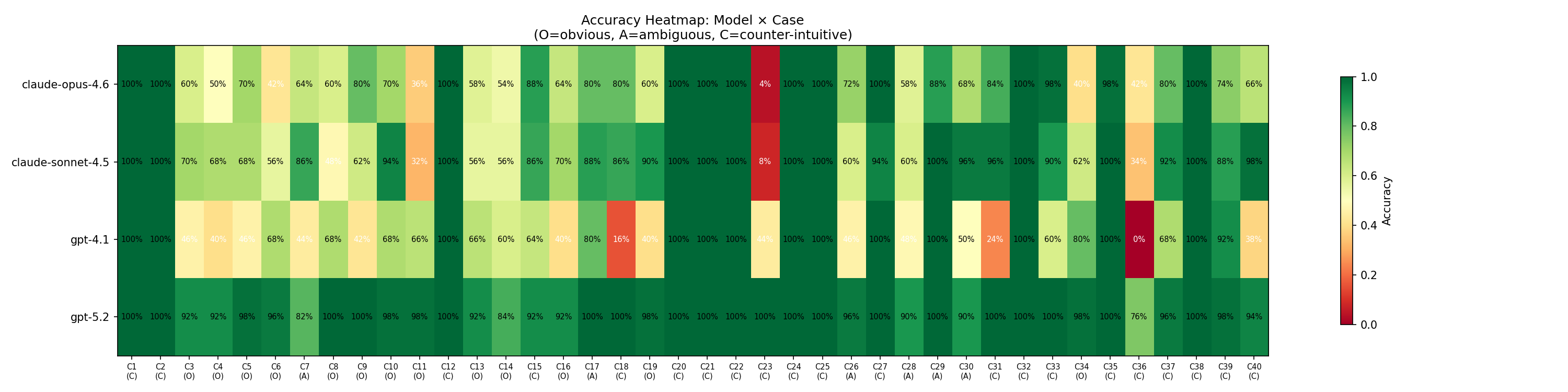}
\caption{Heatmap of accuracy by model and case across all prompt strategies and repetitions. Red cells indicate systematic failures; cases are sorted by intuitiveness. Notable failure clusters appear on Cases~23 (daycare fines) and 36 (right-to-carry laws).}
\label{fig:heatmap}
\end{figure*}

Figure~\ref{fig:heatmap} reveals the full pattern: while most cases are solved reliably by at least one model, the hardest counter-intuitive cases produce systematic failures across all models.

\subsection{Familiarity as a Non-Factor}
\label{sec:familiarity}

Adding log(citations) as a moderator yields a non-significant effect (OR = \statORfam{}, $p =$ \statPfam{}). Models are not simply ``looking up'' well-known results: Case~23 (daycare late pickup fines), based on one of the most widely cited behavioral economics experiments \citep{gneezyRustichini2000}, achieves only \statAccCaseDaycare{}\% accuracy. This speaks to the ``causal parrots'' debate \citep{zevcevic2023causal}: if LLMs merely recited memorized facts, accuracy should correlate with citation frequency. The null result suggests models have absorbed causal knowledge broadly, but deploying it correctly requires a reasoning step that fails when intuitions conflict.

% ============================================================
\section{Discussion}
\label{sec:discussion}
% ============================================================

\subsection{Slow Talking, Not Slow Thinking}

Our CoT paradox challenges the assumption that CoT elicits deeper reasoning. On counter-intuitive cases, the interaction OR of \statORCotXCI{} attenuates CoT's main-effect log-odds benefit by \statAttenuationPct{}\% (from $\ln(\statORcotMain{}) = \statLnORcotMain{}$ to an effective $\ln(\statEffectiveORCotCI{}) = \statLnEffectiveORCotCI{}$). CoT still helps---the effective OR of $\approx$\statEffectiveORCotCI{} remains significant---but the attenuation is substantial and highly significant ($p$ \statPCotXCI{}). The reasoning chain's first step appears anchored by the same intuitive prior that produces errors, and subsequent steps partially---but not fully---overcome this anchor. This is consistent with \citet{jacoviACL2024cot}, who show that reasoning chains are only as strong as their weakest step.

From the dual-process perspective \citep{kahneman2011thinking}, effective System~2 reasoning requires \emph{inhibiting} System~1. Our results suggest that LLMs' CoT achieves \emph{partial} inhibition---enough to help on many cases, but insufficient when intuitive priors are strong. The result is ``slow talking'' that only sometimes constitutes genuine ``slow thinking.''

\subsection{Implications for Policy Applications}

The overall accuracy of \statAccOverall{}\% is inflated by the one-sided test format (\statAccOneSided{}\% accuracy), where the binary YES/NO choice simplifies the task. On two-sided cases---where careful directional prediction matters most---accuracy drops to \statAccTwoSided{}\%, and specific counter-intuitive cases fall as low as \statAccCaseRTC{}\% (right-to-carry) and \statAccCaseDaycare{}\% (daycare fines). Policymakers relying on LLM-assisted analysis may receive confident but incorrect guidance precisely where it is most consequential.

% ============================================================
\section{Conclusion}
\label{sec:conclusion}
% ============================================================

We introduce a benchmark for evaluating LLM counterfactual reasoning in policy evaluation, organized around an \emph{intuitiveness} dimension inspired by dual-process theory. The CoT paradox reveals that chain-of-thought prompting helps primarily when the correct answer aligns with intuition; its benefit is substantially attenuated on counter-intuitive cases. Combined with the failure of self-consistency and the null effect of familiarity, this points to a fundamental limitation: current LLMs can retrieve causal knowledge and partially reason with it, but their deliberative processes are insufficient to fully override strong intuitive priors. Improving counterfactual reasoning may require innovations that enable more robust inhibitory control---the ability to recognize when one's first instinct is wrong and to sustain reasoning past the initial anchor.

% ============================================================
\section*{Limitations}
% ============================================================

\paragraph{Benchmark Scale.} Our benchmark comprises \statNcases{} cases---sufficient for mixed-effects modeling with case-level random intercepts (ICC = \statICC{}), but limited in breadth relative to the space of possible policy evaluations. While we cover 10 policy domains across 11 countries, some domains (e.g., crime, education) are more heavily represented than others. Expanding the benchmark would improve generalizability but is constrained by the requirement for established causal evidence.

\paragraph{Intuitiveness Classification.} Our primary intuitiveness classification was performed by the author and validated against human annotator accuracy (3 annotators, Fleiss' $\kappa$ = 0.32, indicating fair agreement). The continuous human accuracy measure (proportion of annotators answering correctly) yields qualitatively identical results in robustness analyses. However, ``intuitiveness'' is inherently subjective and may vary by population---what is counter-intuitive to a layperson may be obvious to a domain expert.

\paragraph{Temporal Validity.} We evaluate models via API at a single point in time (May 2026). Model performance may change with updates, RLHF iterations, or expanded training data. Our results represent a snapshot of current capabilities rather than a permanent characterization.

\paragraph{Prompt Coverage.} We test five prompting strategies, but the space of possible prompts is infinite. Techniques such as self-debate, reflection, or multi-agent discussion might yield different patterns. Our prompts are designed to probe specific theoretical mechanisms (System~1 vs.~System~2 activation) rather than to exhaustively optimize accuracy.

\paragraph{Ground Truth.} While all cases are grounded in peer-reviewed research, some findings remain debated within their fields (e.g., the effect of right-to-carry laws on violent crime). We use the majority finding from the most methodologically rigorous studies, but acknowledge that some ``ground truths'' may be revised by future research.

% ============================================================
\section*{Ethics Statement}
% ============================================================

This study evaluates proprietary LLMs via their public API interfaces. Human annotators provided factual judgments on published research findings for validation purposes only; this does not constitute human subjects research. The benchmark cases are drawn entirely from published, peer-reviewed research in economics and social science; no private or sensitive data were collected or used. All experimental prompts and model outputs are stored locally and do not contain personally identifiable information.

We acknowledge that LLM-assisted policy analysis carries risks if deployed without appropriate scrutiny. Our finding that LLMs systematically fail on counter-intuitive cases---precisely where careful analysis is most valuable---underscores the importance of human oversight in policy-relevant applications. We caution against using LLM outputs as a substitute for rigorous empirical analysis and expert judgment in policy evaluation.

\section*{Acknowledgements}

GitHub Copilot was used for code implementation (experimental scripts, statistical analysis pipelines) and for polishing paper text (grammar and style editing). All research ideas, scientific claims, analyses, and interpretations were conceived and verified by the author.

% ============================================================
% References
% ============================================================

\bibliography{references}

% ============================================================
\appendix
\section{Full Case List}
\label{app:cases}

Table~\ref{tab:full_cases} presents the complete list of \statNcases{} benchmark cases. Each case is grounded in at least one peer-reviewed publication using a credible causal identification strategy. Intuitiveness classifications are validated against human annotator accuracy (\S\ref{sec:intuitiveness}).

\begin{table*}[t]
\centering
\small
\setlength{\tabcolsep}{3pt}
\begin{tabular}{@{}rllllll@{}}
\toprule
\textbf{\#} & \textbf{Policy} & \textbf{Country} & \textbf{Domain} & \textbf{Type} & \textbf{Intuit.} & \textbf{GT} \\
\midrule
1 & NJ Minimum Wage Increase (1992) & US & Labor & 1-sided & CI & NO \\
2 & Oregon Medicaid Expansion (2008) & US & Health & 1-sided & CI & NO \\
3 & EITC Expansion (1986) & US & Labor & 2-sided & Obvious & DECREASE \\
4 & TN STAR Class Size (1985) & US & Education & 2-sided & Obvious & DECREASE \\
5 & Head Start Preschool (1965) & US & Education & 2-sided & Obvious & INCREASE \\
6 & Mexico PROGRESA (1997) & Mexico & Education & 2-sided & Obvious & DECREASE \\
7 & German Reunification (1990) & Germany & Macro & 2-sided & Ambig. & INCREASE \\
8 & CA Tobacco Control (1988) & US & Health & 2-sided & Obvious & INCREASE \\
9 & China WTO Accession (2001) & US/China & Trade & 2-sided & Obvious & INCREASE \\
10 & Recession \& Air Pollution (1981) & US & Health & 2-sided & Obvious & INCREASE \\
11 & Australia Gun Buyback (1996) & Australia & Crime & 2-sided & Obvious & INCREASE \\
12 & UK National Minimum Wage (1999) & UK & Labor & 1-sided & CI & NO \\
13 & India NREGA (2006) & India & Labor & 2-sided & Obvious & DECREASE \\
14 & Colombia Health Insurance (1993) & Colombia & Health & 2-sided & Obvious & DECREASE \\
15 & Romania Abortion Ban (1966) & Romania & Education & 2-sided & CI & INCREASE \\
16 & MTO Housing Vouchers (1990s) & US & Social & 2-sided & Obvious & DECREASE \\
17 & Abortion Legalization \& Crime & US & Crime & 2-sided & Ambig. & INCREASE \\
18 & Ban the Box (2000s) & US & Labor & 2-sided & CI & INCREASE \\
19 & Israel Class Size Rule & Israel & Education & 2-sided & Obvious & DECREASE \\
20 & Cash for Clunkers (2009) & US & Macro & 1-sided & CI & NO \\
21 & Microfinance (6 countries) & Multi & Development & 1-sided & CI & NO \\
22 & Mariel Boatlift (1980) & US & Labor & 1-sided & CI & NO \\
23 & Daycare Late Pickup Fines (1998) & Israel & Behavioral & 2-sided & CI & DECREASE \\
24 & Death Penalty \& Murder Rate & US & Crime & 1-sided & CI & NO \\
25 & Medical Marijuana \& Teen Use & US & Health & 1-sided & CI & NO \\
26 & TANF Welfare Reform (1996) & US & Social & 2-sided & Ambig. & DECREASE \\
27 & Foreign Aid \& GDP Growth & Multi & Development & 1-sided & CI & NO \\
28 & No-Excuses Charter Schools & US & Education & 2-sided & Ambig. & DECREASE \\
29 & Grade Retention & US & Education & 1-sided & Ambig. & NO \\
30 & Police Staffing \& Crime & US & Crime & 2-sided & Ambig. & INCREASE \\
31 & Scared Straight Programs & US & Crime & 2-sided & CI & DECREASE \\
32 & DARE Anti-Drug Program & US & Health & 1-sided & CI & NO \\
33 & Job Corps Training & US & Labor & 1-sided & CI & NO \\
34 & Kenya Deworming (1998) & Kenya & Health & 2-sided & Obvious & DECREASE \\
35 & Meth Precursor Controls (1995) & US & Crime & 1-sided & CI & NO \\
36 & Right-to-Carry Laws (1977--2014) & US & Crime & 2-sided & CI & DECREASE \\
37 & SF Rent Control (1994) & US & Housing & 2-sided & CI & INCREASE \\
38 & Abstinence-Only Education & US & Health & 1-sided & CI & NO \\
39 & China One-Child Policy \& Savings & China & Macro & 2-sided & CI & DECREASE \\
40 & Violent Movies \& Crime & US & Crime & 2-sided & CI & INCREASE \\
\bottomrule
\end{tabular}
\caption{Complete list of \statNcases{} benchmark cases. Type: 1-sided = YES/NO, 2-sided = directional. Intuit.: CI = counter-intuitive, Ambig.\ = ambiguous. GT = ground truth answer.}
\label{tab:full_cases}
\end{table*}

\section{Prompt Templates}
\label{app:prompts}

We use five prompting strategies, each with variants for one-sided (YES/NO) and two-sided (INCREASE/DECREASE/NO SIGNIFICANT CHANGE) test formats. Template variables: \texttt{\{policy\}}, \texttt{\{country\}}, \texttt{\{year\}}, \texttt{\{outcome\}}, \texttt{\{hypothesis\}} (one-sided only).

\subsection*{P1 --- Na\"ive (Two-sided)}
\begin{quote}
\small
Consider the following policy: \{policy\}, implemented in \{country\} in \{year\}. If this policy had NOT been implemented, what would have happened to \{outcome\}? Please answer with exactly one of: INCREASE, DECREASE, or NO SIGNIFICANT CHANGE. Answer:
\end{quote}

\subsection*{P2 --- Expert Persona (Two-sided)}
\begin{quote}
\small
You are an experienced econometrician specializing in policy evaluation. You are familiar with difference-in-differences, synthetic control, regression discontinuity, and instrumental variable methods. Consider the following policy: \{policy\}, implemented in \{country\} in \{year\}. Based on your understanding of the causal mechanisms involved, if this policy had NOT been implemented, what would have happened to \{outcome\}? Please answer with exactly one of: INCREASE, DECREASE, or NO SIGNIFICANT CHANGE. Answer:
\end{quote}

\subsection*{P3 --- Chain-of-Thought (Two-sided)}
\begin{quote}
\small
Consider the following policy: \{policy\}, implemented in \{country\} in \{year\}. If this policy had NOT been implemented, what would have happened to \{outcome\}? Think step by step: 1.\ What are the main causal channels through which this policy affects \{outcome\}? 2.\ Are there any confounding factors or alternative explanations? 3.\ Based on your analysis, what is the most likely counterfactual direction? Final answer (exactly one of: INCREASE, DECREASE, or NO SIGNIFICANT CHANGE):
\end{quote}

\subsection*{P4 --- Structured Reasoning (Two-sided)}
\begin{quote}
\small
Consider the following policy: \{policy\}, implemented in \{country\} in \{year\}. I want you to analyze the counterfactual: what would have happened to \{outcome\} if this policy had NOT been implemented? Please structure your response as follows: Causal mechanisms: List the main channels. Confounders: List any confounding factors. Counterfactual prediction: The most likely direction. Final answer (exactly one of: INCREASE, DECREASE, or NO SIGNIFICANT CHANGE):
\end{quote}

\subsection*{P5 --- Adversarial Framing (Two-sided)}
\begin{quote}
\small
Consider the following policy: \{policy\}, implemented in \{country\} in \{year\}. Some analysts argue that this policy had NO meaningful effect on \{outcome\}---that the observed changes were driven entirely by other factors such as pre-existing trends, concurrent policies, or macroeconomic conditions. Do you agree with this assessment? If this policy had NOT been implemented, would \{outcome\} have been essentially the same? Please answer with exactly one of: INCREASE, DECREASE, or NO SIGNIFICANT CHANGE. Answer:
\end{quote}

\noindent One-sided variants follow the same structure but replace the outcome question with the pre-formulated \texttt{\{hypothesis\}} and use YES/NO as response options.

\section{Additional Results}
\label{app:additional}

\begin{figure}[t]
\centering
\includegraphics[width=\columnwidth]{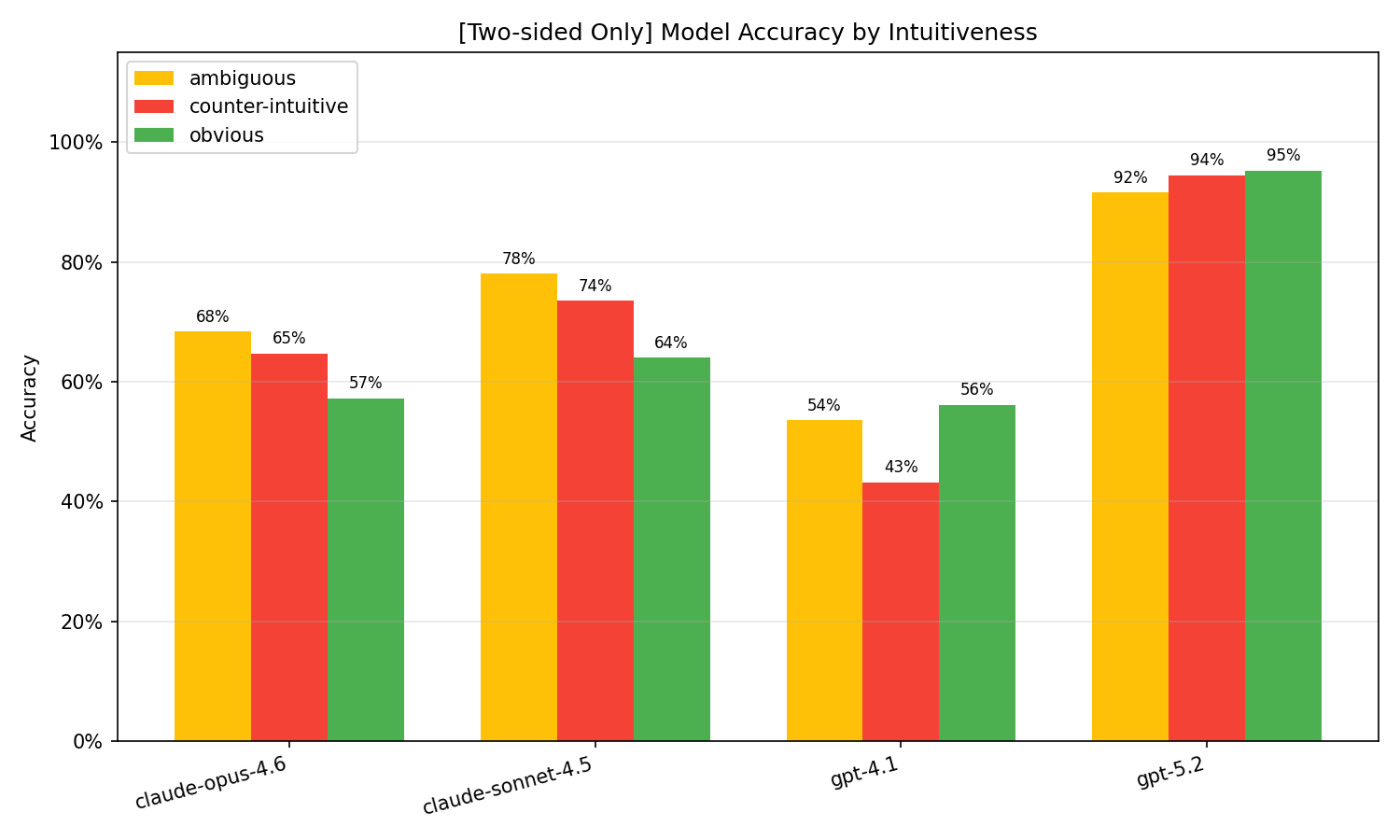}
\caption{Accuracy by model and intuitiveness category on two-sided cases. GPT-5.2 maintains high accuracy across categories, while GPT-4.1 drops to \statAccModelGptFourTsCI{}\% on counter-intuitive cases.}
\label{fig:model_intuitiveness}
\end{figure}

\begin{figure}[t]
\centering
\includegraphics[width=\columnwidth]{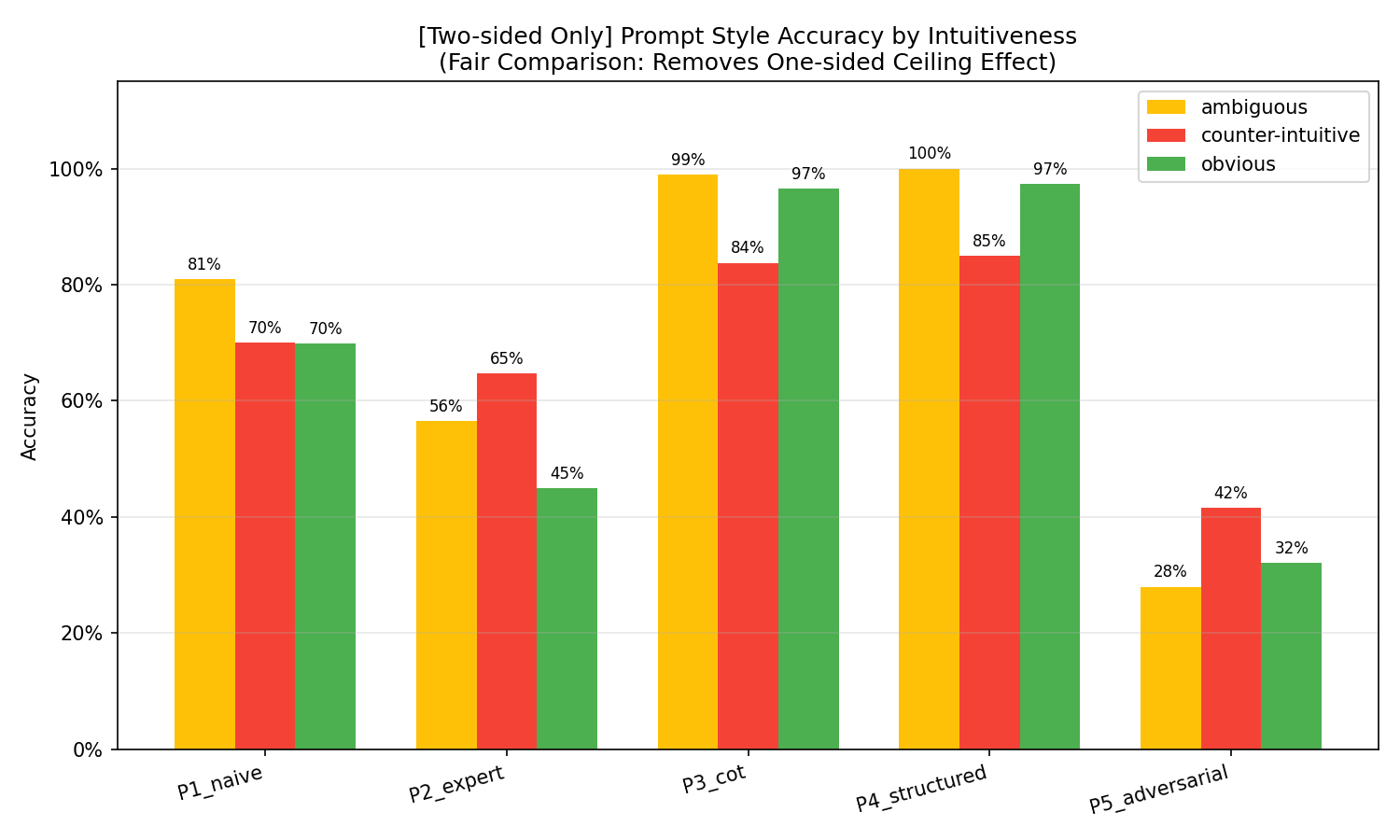}
\caption{Accuracy by prompt strategy and intuitiveness category on two-sided cases. CoT and structured prompts show uniformly high accuracy on obvious cases, but their advantage diminishes on counter-intuitive cases.}
\label{fig:prompt_intuitiveness}
\end{figure}

\begin{figure}[t]
\centering
\includegraphics[width=\columnwidth]{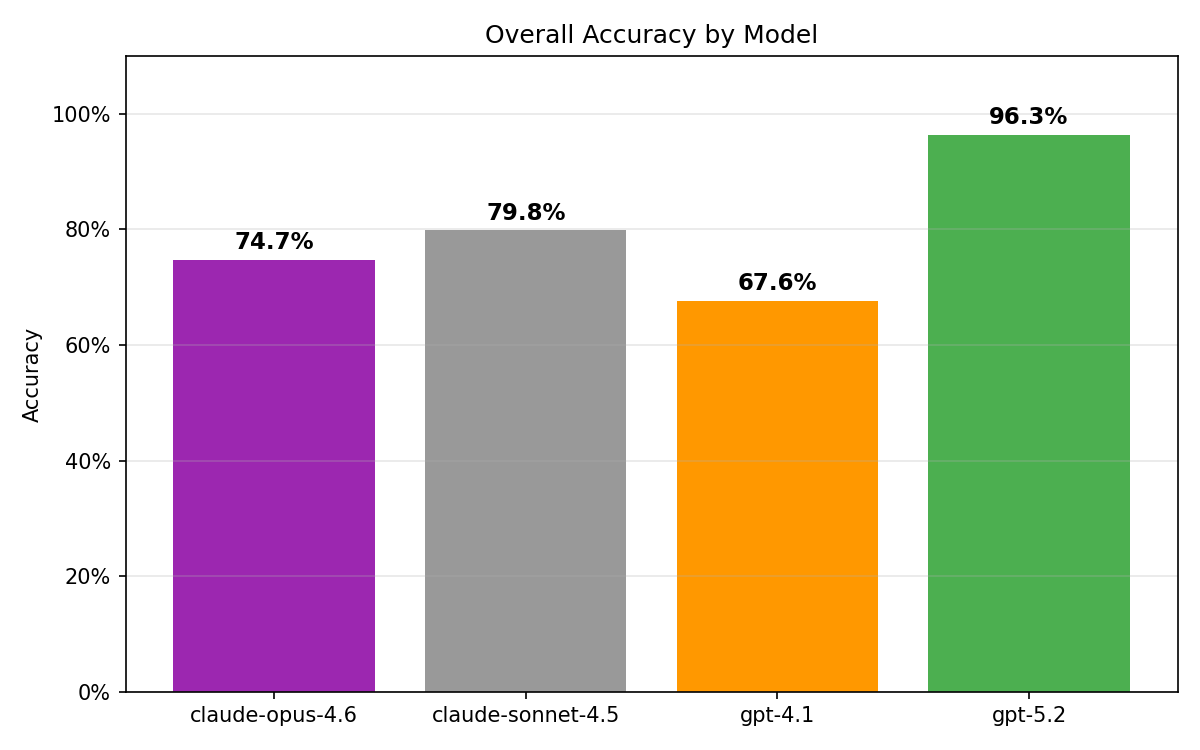}
\caption{Overall accuracy by model across all conditions. Error bars show 95\% binomial confidence intervals.}
\label{fig:model_accuracy_all}
\end{figure}

\begin{figure}[t]
\centering
\includegraphics[width=\columnwidth]{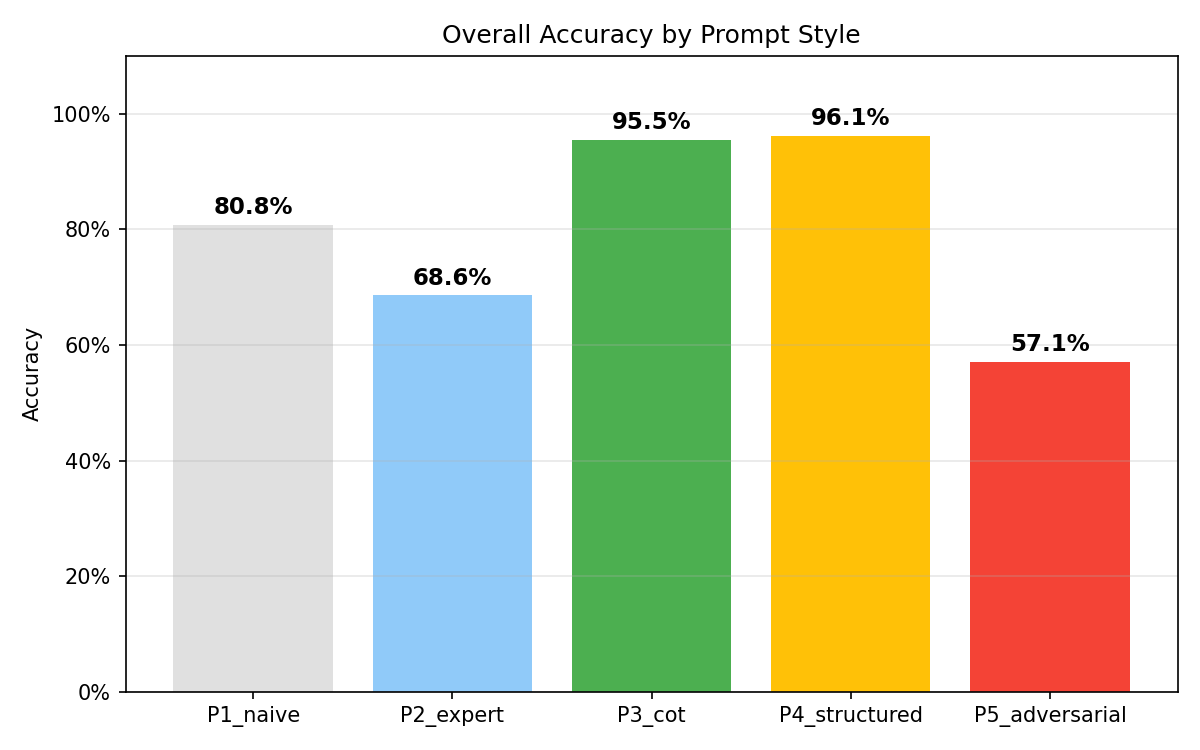}
\caption{Overall accuracy by prompt strategy across all conditions. Structured reasoning (P4) and CoT (P3) substantially outperform na\"ive (P1), expert (P2), and adversarial (P5) prompts.}
\label{fig:prompt_accuracy_all}
\end{figure}

\section{Human Annotation Details}
\label{app:annotation}

Three annotators with diverse academic backgrounds (economics, statistics, medicine) independently predicted the outcome direction for all \statNcases{} cases without access to the answers. Table~\ref{tab:annotators} summarizes annotator characteristics.

\begin{table}[t]
\centering
\small
\begin{tabular}{@{}llrr@{}}
\toprule
\textbf{ID} & \textbf{Background} & \textbf{Familiarity} & \textbf{Accuracy} \\
\midrule
A1 & Economics & 5/5 & 26/40 (65\%) \\
A2 & Statistics & 3/5 & 25/40 (63\%) \\
A3 & Medicine & 2/5 & 25/40 (63\%) \\
\midrule
\multicolumn{2}{l}{Fleiss' $\kappa$} & \multicolumn{2}{r}{0.32 (fair)} \\
\bottomrule
\end{tabular}
\caption{Human annotator characteristics. Familiarity = self-rated familiarity with economics/social science (1--5). The moderate $\kappa$ reflects genuine disagreement on counter-intuitive cases, consistent with our task design.}
\label{tab:annotators}
\end{table}

The continuous intuitiveness measure (\texttt{human\_acc} = proportion of annotators answering correctly) takes values in $\{0, 1/3, 2/3, 1\}$ and is used as a robustness check for the categorical classification (see \S\ref{sec:intuitiveness}).

\end{document}